# An ERP Study on Recursive Locative Processing in Mandarin-Speaking Children with Autism

Wang Xiaoyi[1] Fu chenxi[2] Zhuang Ziman[3] Yang Caimei[4]

## Abstract

Recursion enables the generation of hierarchical linguistic structures but imposes substantial processing demands during real-time comprehension. While difficulties with complex syntax have been reported in autism spectrum disorder (ASD), the temporal dynamics of recursive processing remain poorly understood.

This study used event-related potentials (ERPs) to examine how Mandarin-speaking children with ASD process two-level recursive locative constructions. Twenty-four children (12 ASD, 12 typically developing, TD) participated in a cross-modal sentence–picture matching task. Neural responses were analyzed across three processing stages associated with structural prediction (P200), semantic integration (N400), and syntactic reanalysis (P600), with mental age controlled.

Results revealed a systematic divergence between groups. TD children showed clear P200 and P600 modulation in response to structural mismatch, whereas ASD children exhibited attenuated early differentiation and reduced late reanalysis effects. In contrast, ASD children showed enhanced N400 responses under mismatch conditions, indicating increased semantic integration demands. In addition, the ASD group displayed significantly greater inter-individual variability in hemispheric lateralization, although lateralization strength was not associated with receptive vocabulary performance.

These findings support a cascading account in which reduced early predictive engagement in ASD leads to increased integration costs and diminished reanalysis efficiency during recursive processing. More broadly, the results highlight the importance of both temporal processing dynamics and neural variability in understanding language differences in ASD.

[1] Wang Xiaoyi is currently studying at Soochow University. Her research area covers child language acquisition and she is specialized in recursive acquisition. E-mail address: wangxiaoyi224@163.com
[2] Fu Chenxi is currently studying at Soochow University. Her research area covers first language acquisition. E-mail address: fuchenxi_edu@outlook.com

## 1. Introduction

Recursion is widely regarded as a defining property of human language within generative grammar. It enables the generation of an infinite set of expressions from a finite system through the repeated application of a computational operation (Chomsky, 2014; Hauser et al., 2002). In the Minimalist Program, this capacity is formalized as Merge, which recursively combines linguistic elements into hierarchical structures (Chomsky et al., 2023). While recursion defines grammatical competence, a substantial body of research suggests that the real-time processing of recursive structures imposes increasing cognitive demands, particularly as hierarchical depth grows. This highlights a well-established distinction between competence and performance, whereby structurally permissible sentences may nonetheless challenge processing systems.

Such processing demands are especially relevant for children with Autism Spectrum Disorder (ASD). Although individuals with ASD may perform comparably to typically developing (TD) peers on simple sentence structures, difficulties emerge when processing hierarchically complex constructions, including recursive embedding and long-distance dependencies (Jacobs & Richdale, 2013; Andreou et al., 2025). These challenges have been linked to limitations in working memory, predictive processing, and contextual integration (Adlof & Catts, 2015; Peristeri & Tsimpli, 2020). Importantly, this difficulty appears to be systematic rather than incidental, suggesting that hierarchical structure itself may place disproportionate demands on the ASD processing system.

To investigate such difficulties, recursive locative constructions provide a particularly suitable testing domain. Structures such as “the bird on the crocodile in the river” instantiate hierarchical embedding through successive locative modification. In Mandarin, this process is realized through the recursive application of the DeP structure, where the functional element 的 mediates hierarchical combination (Mao et al., 2024). These constructions require the mapping of linear input onto hierarchical representations, placing direct demands on real-time processing. As such, they offer a controlled way to examine how increasing structural depth affects comprehension.

Despite growing interest in recursion and ASD, several gaps remain. Existing research has primarily relied on behavioral measures, which capture end-state performance but do not reveal the temporal dynamics of processing. Event-related potentials (ERPs), by contrast, provide millisecond-level resolution and allow the dissociation of processing stages, including early prediction (P200), semantic integration (N400), and syntactic reanalysis (P600) (Kutas & Federmeier, 2011; Osterhout & Holcomb, 1992). However, ERP studies on recursive processing in ASD remain limited, particularly for Mandarin-speaking children and for locative recursive structures.

The present study addresses this gap by examining the neural dynamics of Mandarin locative recursive processing in children with ASD compared to TD peers. Specifically, it investigates (1) whether early and mid-stage ERP responses (P200, N400) differ between groups, (2) whether late syntactic reanalysis processes (P600) are differentially modulated, and (3) whether hemispheric lateralization patterns vary across individuals and relate to receptive vocabulary ability.

## 2. Theoretical Background

### 2.1 Recursion and Locative Structures

Recursion refers to the hierarchical embedding of a syntactic category within itself, yielding structures of potentially unbounded depth (Chomsky et al., 2002). A key distinction is drawn between recursive embedding and iteration, the latter involving linear repetition without increasing hierarchical complexity (Rohrmeier et al., 2014). Within natural language, recursive structures frequently occur in possessive and locative constructions, often instantiated as tail recursion (Yang, 2014).

Recursive locative constructions involve successive embedding of locative modifiers, forming hierarchical representations of spatial relations. In Mandarin, this process is typically realized through DeP structures, in which the functional element 的 mediates recursive modification (Mao et al., 2024). These constructions require the processing system to maintain hierarchical representations while incrementally integrating incoming linguistic input, thereby placing increasing cognitive demands as embedding depth increases.

Cross-linguistic research indicates that recursive structures are acquired in a stage-wise manner, with children demonstrating higher accuracy for single-level embedding than for deeper hierarchical structures. However, developmental trajectories vary across languages and tasks, suggesting that both linguistic properties and methodological factors influence performance (Pérez-Leroux et al., 2023; Mao et al., 2024). Overall, increasing embedding depth consistently imposes greater processing demands, particularly in populations with limited cognitive resources.

**2.2 Predictive Processing in Language Comprehension**

Prediction is widely considered a core mechanism in language comprehension. Contemporary accounts conceptualize prediction as a probabilistic, multi-level process in which the comprehender generates and updates expectations about upcoming linguistic input (Kuperberg & Jaeger, 2016). During comprehension, multiple hypotheses are maintained and continuously revised as new input is encountered.

In ASD, atypical predictive processing has been proposed as a key underlying mechanism (Sinha et al., 2014; Van de Cruys et al., 2014). Within language processing, such differences may manifest as reduced efficiency in integrating incoming information with prior expectations, particularly in structurally complex contexts. Recursive constructions, by increasing hierarchical depth, place growing demands on prediction and integration processes, making them an informative domain for examining these mechanisms.

ERP components provide a useful means of investigating these processes. The N400 is typically associated with semantic integration and prediction-related processing, whereas the P600 has been linked to syntactic reanalysis and structural updating (Kutas & Federmeier, 2011). Variations in these components may therefore reflect differences in how predictions are generated and updated during recursive processing.

### 2.3 Hemispheric Lateralization

Language processing is typically associated with left-hemisphere dominance, although this pattern emerges gradually during development and exhibits interindividual variability (Somers et al., 2015; Ocklenburg & Güntürkün, 2017). Early in development, language processing often involves more bilateral engagement, with increasing specialization over time.

In ASD populations, atypical patterns of lateralization have been reported, including reduced leftward dominance and increased bilateral or right-hemisphere involvement (Lindell, 2020; Jouravlev et al., 2020). Such differences may be particularly pronounced when processing hierarchically complex linguistic structures.

Recursive locative constructions provide a suitable domain for examining lateralization, as they require the integration of hierarchical syntactic information. By analyzing the topographic distribution of ERP components, the present study investigates whether hemispheric engagement differs between ASD and TD groups during recursive processing.

### 2.4 Research Gap

Existing ERP research suggests that individuals with ASD often exhibit altered neural responses during language processing, including delayed or attenuated N400 and P600 effects, particularly in contexts involving structural complexity (Koolen et al., 2014; Maia et al., 2021; Márquez-García et al., 2022). However, several limitations remain.

First, relatively few studies have examined recursive structures in ASD, and even fewer have focused on locative recursion. Second, ERP investigations of recursive processing in Mandarin-speaking children are limited. Third, variability within the ASD population has not been systematically examined in relation to recursive processing.

The present study addresses these gaps by investigating the neural dynamics of Mandarin locative recursion in children with ASD, using ERP measures to examine both group-level differences and individual variability.

# 3. Methodology

## 3.1 Participants

Twenty-four Mandarin-speaking children participated in the study, including 12 children with Autism Spectrum Disorder (ASD) and 12 typically developing (TD) children. ASD participants were recruited from specialized education institutions in Suzhou, while TD children were recruited from a mainstream kindergarten in the same city. All participants were right-handed, with normal hearing and vision, and no additional neurological or psychiatric conditions.

Table 1 Subject Information

| | Number | Average age | Average mental age | Average PPVT score |
|---|---|---|---|---|
| ASD | 12 | 6.36 | 5.58 | 70.36 |
| TD | 12 | 6.32 | 7.9 | 123.75 |

Prior to participation, written informed consent was obtained from guardians. The study was approved through an institutional ethical review process. To ensure comparability, all participants completed standardized assessments: the Wechsler Intelligence Scale for Children (WISC) for cognitive ability and the Peabody Picture Vocabulary Test (PPVT) for receptive language. ASD diagnoses were confirmed via institutional records and clinical certificates. Participants were approximately six years old (ASD: M = 6.36; TD: M = 6.32). This age range was selected based on prior research identifying age seven as a critical stage in the development of recursive structures.

## 3.2 Research Questions

This study investigates the neural processing of recursive locative structures in Mandarin-speaking children with ASD using event-related potentials (ERPs). The following questions were addressed:

Do ASD and TD children differ in early (P200) and semantic (N400) ERP responses during recursive processing?

Are there group differences in P600 (500–800 ms), reflecting syntactic integration or reanalysis?

Do children with ASD exhibit greater variability in hemispheric lateralization (LI), and is this variability related to language ability (PPVT)?

**3.3 Experimental Design and Procedure**

A picture–sentence verification paradigm was employed. On each trial, participants viewed an image and ןutened to a sentence, then judged whether the sentence matched the image via button press. This design minimizes movement artifacts and is suitable for EEG recording in children.

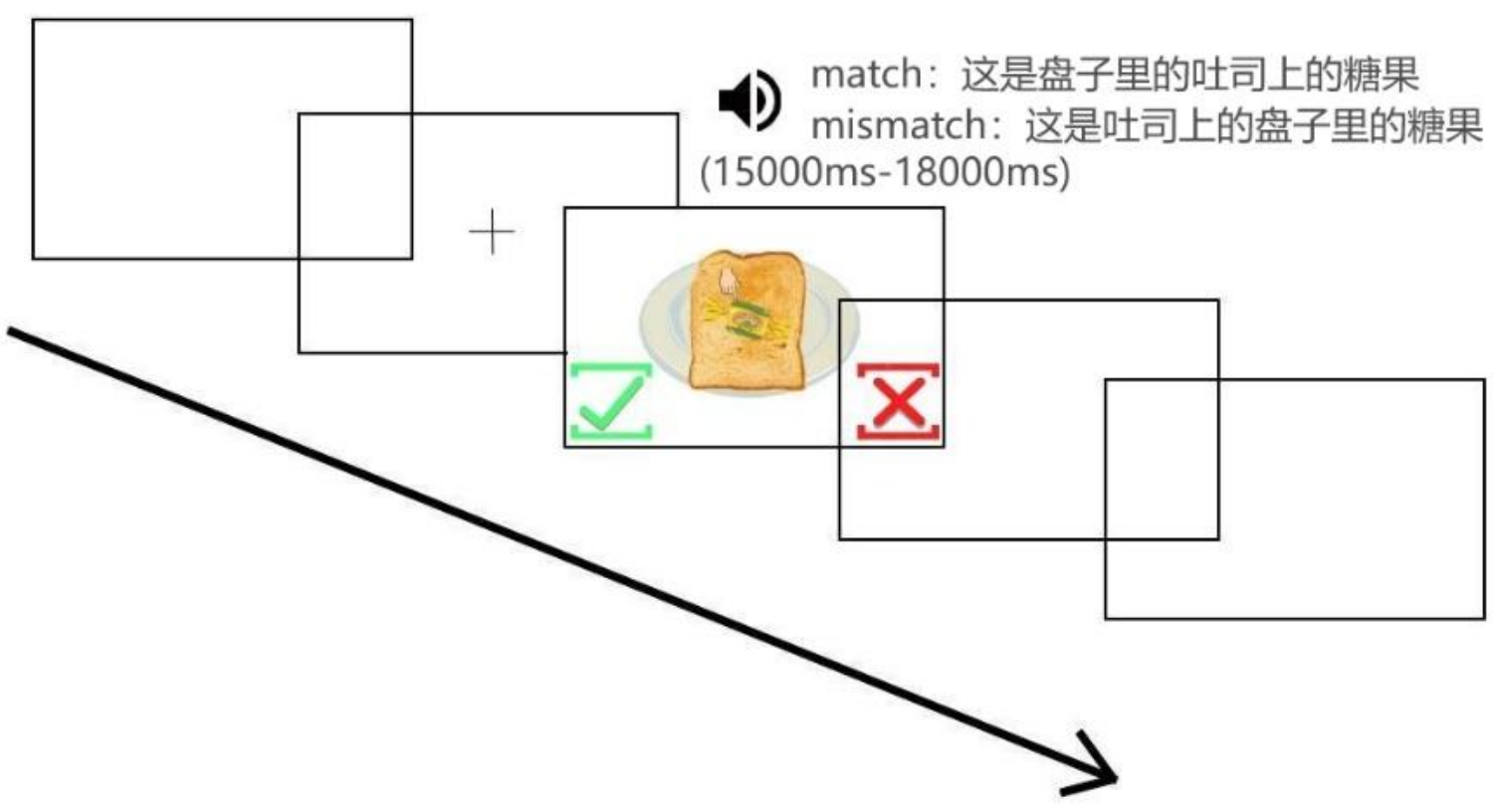

Figure 1 Example Stimuli of ERP Sentence-picture Matching Task

Stimuli depicted two-level recursive locative structures. Each image was paired with both a matching and a mismatching sentence (e.g., correct vs. reversed locative order), yielding 120 trials (60 match, 60 mismatch). Sentences were presented word-by-word using text-to-speech with a 100 ms inter-word interval to ensure precise temporal alignment.

Critical time-locking points included the onset of the second locative phrase (LocP2) and the final noun. LocP2 was used to examine early predictive processing (P200), while the noun onset indexed semantic and syntactic integration (N400, P600).

Trials were divided into four randomized blocks. Participants completed a practice session (10 trials) and were required to reach 75% accuracy before proceeding. The experiment was programmed in E-Prime 3.0.

Table 2 The Examples of Experiment Materials

| Group | Condition | Picture | Audio (word form) |
|---|---|---|---|
| ASD | Match | | 这是魔方上的尺子上的花朵<br>zhè shì mofang shàng de chizi shàng de huaduo.<br>This is the flower on the ruler on the Rubik's Cube. |
| | Mismatch | | 这是尺子上的魔方上的花朵<br>zhè shì chizi shàng de mofang shàng de huaduo.<br>This is the flower on the Rubik's Cube on the ruler |
| TD | Match | | 这是魔方上的尺子上的花朵<br>zhè shì mofang shàng de chizi shàng de huaduo.<br>This is the flower on the ruler on the Rubik's Cube. |
| | Mismatch | | 这是尺子上的魔方上的花朵<br>zhè shì chizi shàng de mofang shàng de huaduo.<br>This is the flower on the Rubik's Cube on the ruler |

### 3.4 EEG Recording and Data Analysis

EEG data were recorded using a 32-channel Smarting PRO system following the 10–20 system. Signals were sampled at 250 Hz with a 50 Hz notch filter applied online. Data were re-referenced offline to TP9/TP10.

Preprocessing was conducted in EEGLAB. Data were filtered (0.1–30 Hz),

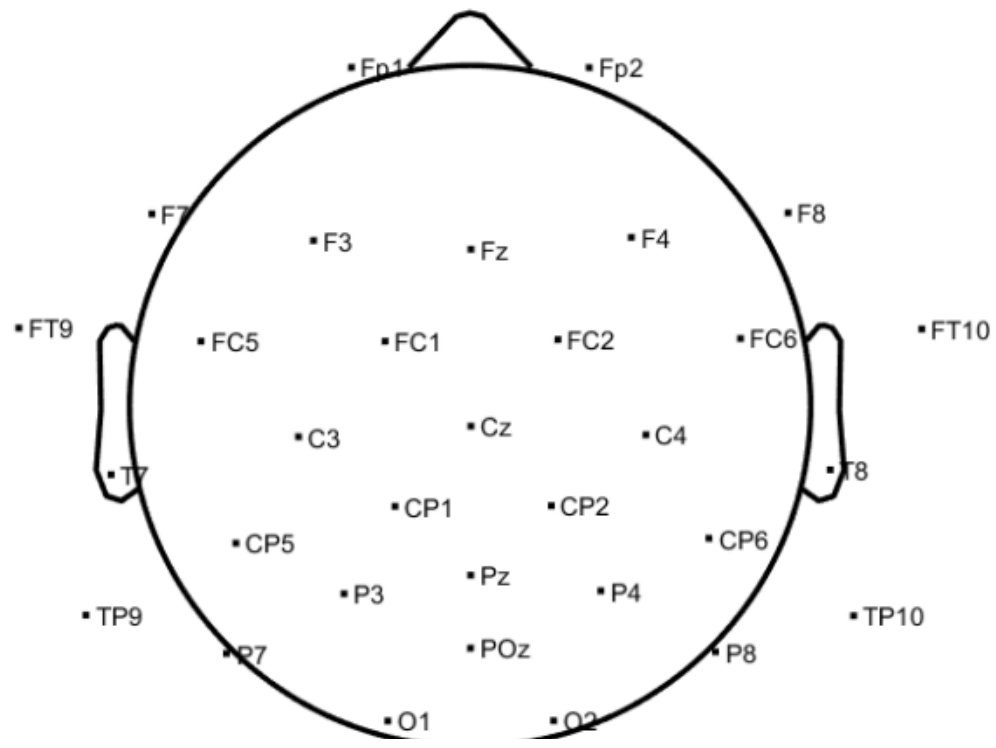

Figure 2 The 2D Diagram of Electrode on Smarting PRO Cap 32

segmented into epochs (−200 to 1000 ms), and baseline-corrected. Artifacts were removed via visual inspection and independent component analysis (ICA). Epochs exceeding ±175 μV were rejected.

ERP analyses focused on mean amplitudes in the P200, N400, and P600 time windows. Electrodes were grouped into four regions of interest (ROIs): left/right × anterior/posterior.

Statistical analyses were conducted using linear mixed-effects models in R. Group (ASD vs. TD) was treated as a between-subject factor, with Condition, Hemisphere, and Region as within-subject factors. Mental age was included as a covariate. Post-hoc comparisons were corrected using Bonferroni adjustment.

**3.5 Lateralization Analysis**

To examine hemispheric asymmetry, a lateralization index (LI) was computed for each participant during the P600 window:

Absolute LI values (|LI|) were analyzed to assess variability in lateralization strength. Group differences were tested using non-parametric methods (Wilcoxon rank-sum; Fligner–Killeen test).

Extreme lateralization cases were identified using the interquartile range criterion and compared using Fisher’s Exact Test. Additionally, exploratory clustering was performed on signed LI values to characterize distributional patterns.

Within the ASD group, robust regression was used to examine the relationship between |LI| and PPVT scores.

# 4. Results

## 4.1 RQ1: P200 and N400 Effects

To examine early processing, mean amplitudes in the 200–350 ms window time-locked to the onset of the second locative phrase were analyzed over bilateral anterior regions. Descriptive statistics showed that the TD group exhibited a larger positivity under mismatch (M = 1.88 μV, SD = 3.44) than match conditions (M = 0.45 μV, SD = 2.16), whereas the ASD group showed no such enhancement (mismatch: M = 0.00 μV, SD = 2.21; match: M = 0.43 μV, SD = 1.82).

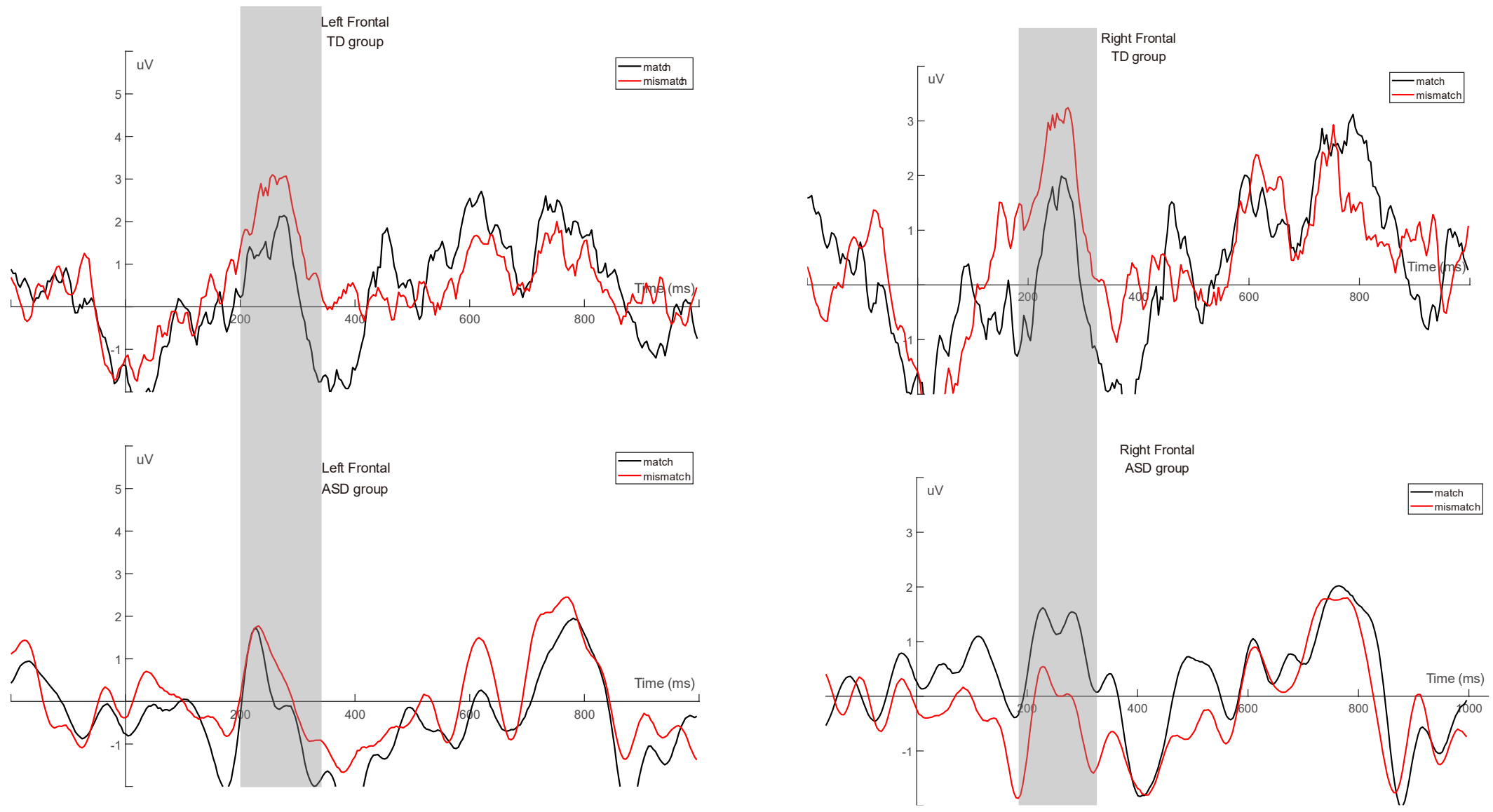

Figure 3 Waveforms of Frontal Areas for Children in ASD and TD Groups

A linear mixed-effects model (LMM; MeanVoltage ~ Group × Condition × Region + MA_c + (1 | Subject)) revealed a significant main effect of Group, $F(1, 390.23) = 72.91$, $p < .001$, and Condition, $F(1, 380.83) = 5.31$, $p = .022$, as well as a significant Group × Condition interaction, $F(1, 380.83) = 18.37$, $p < .001$. Mental age was also significant, $F(1, 390.08) = 61.22$, $p < .001$. Post-hoc comparisons showed that ASD children exhibited reduced amplitudes relative to TD children under both match (estimate = −1.68 μV, $p < .001$) and mismatch conditions (estimate = −3.55 μV, $p < .001$), with a larger group difference in the mismatch condition.

Table 3 Descriptive Statistics of P200 Mean Amplitude (μV) for Each Group and Condition.

| Group | Condition | n | Mean | SD |
|---|---|---|---|---|
| ASD | match | 12 | 0.43 | 1.82 |
| | mismatch | 12 | 0 | 2.21 |
| TD | match | 12 | 0.45 | 2.16 |
| | mismatch | 12 | 1.88 | 3.44 |

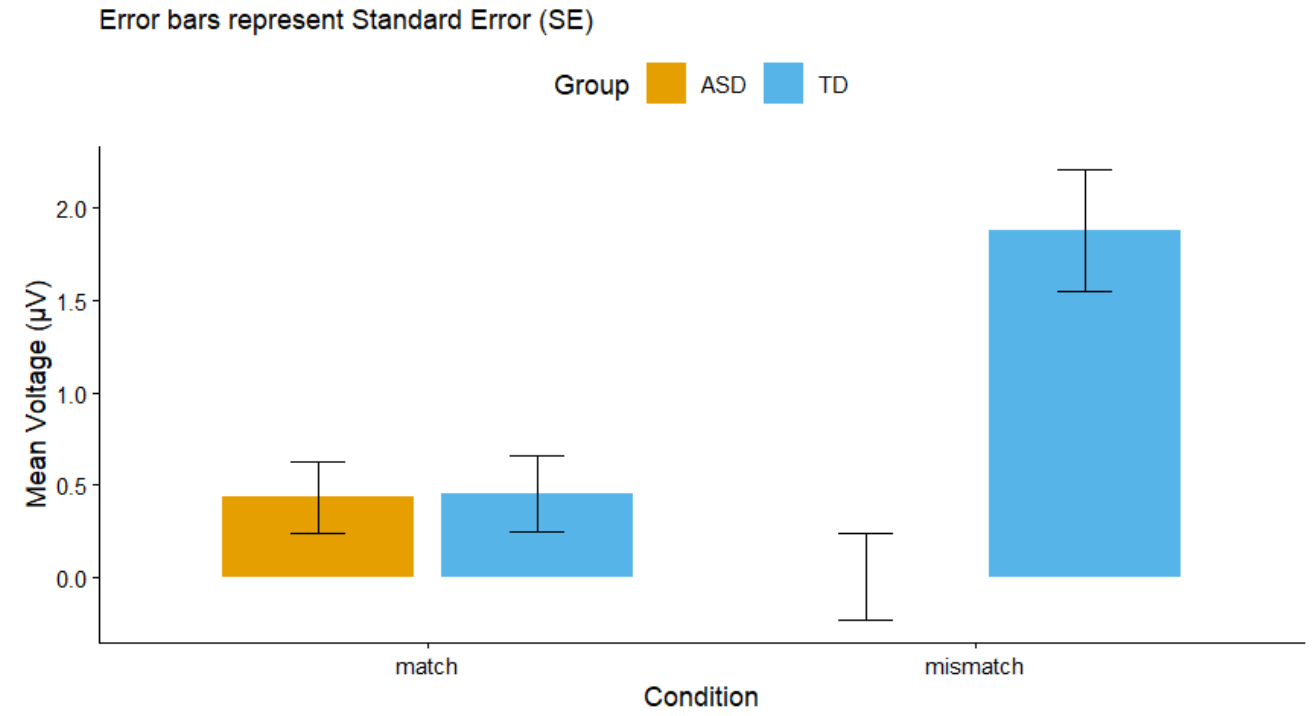

Figure 4 P200 Mean Voltage by Group and Condition

For the N400 (300–500 ms, time-locked to the final noun), analyses were conducted across four ROIs (LF, RF, LP, RP). Descriptive statistics indicated that the ASD group showed a stronger negativity under mismatch (M = −1.59 μV, SD = 1.65) compared to match (M = −0.66 μV, SD = 2.72), whereas the TD group showed minimal condition differences.

The LMM revealed a significant main effect of Condition, F(1, 692.81) = 7.81, p = .005, and Region, F(3, 692.81) = 8.12, p < .001, but no main effect of Group, F(1, 702.58) = 0.84, p = .358. Crucially, the Group × Condition interaction was significant, F(1, 692.81) = 10.11, p = .002. Post-hoc comparisons showed no group difference under match (p = .276), but significantly greater negativity in the ASD group under mismatch (estimate = −0.72 μV, p = .010).

Table 4 Fixed Effects Results of the Linear Mixed-effects Model for P200 Amplitude

| Effect) | df | F value | p |
|---|---|---|---|
| Group | 1, 390.2 | 72.91 | < .001 *** |
| Condition | 1, 380.8 | 5.31 | .022 * |
| MA | 1, 390.1 | 61.22 | < .001 *** |
| Group × Condition | 1, 380.8 | 18.37 | < .001 *** |

Table 5 Simple Effect Analysis of the Group × Condition Interaction.

| Condition | Contrast | Estimate | SE | t | P |
|---|---|---|---|---|---|
| Match | ASD - TD | -1.68 | 0.38 | -4.47 | < .001 |
| Mismatch | ASD - TD | -3.55 | 0.38 | -9.42 | < .001 |

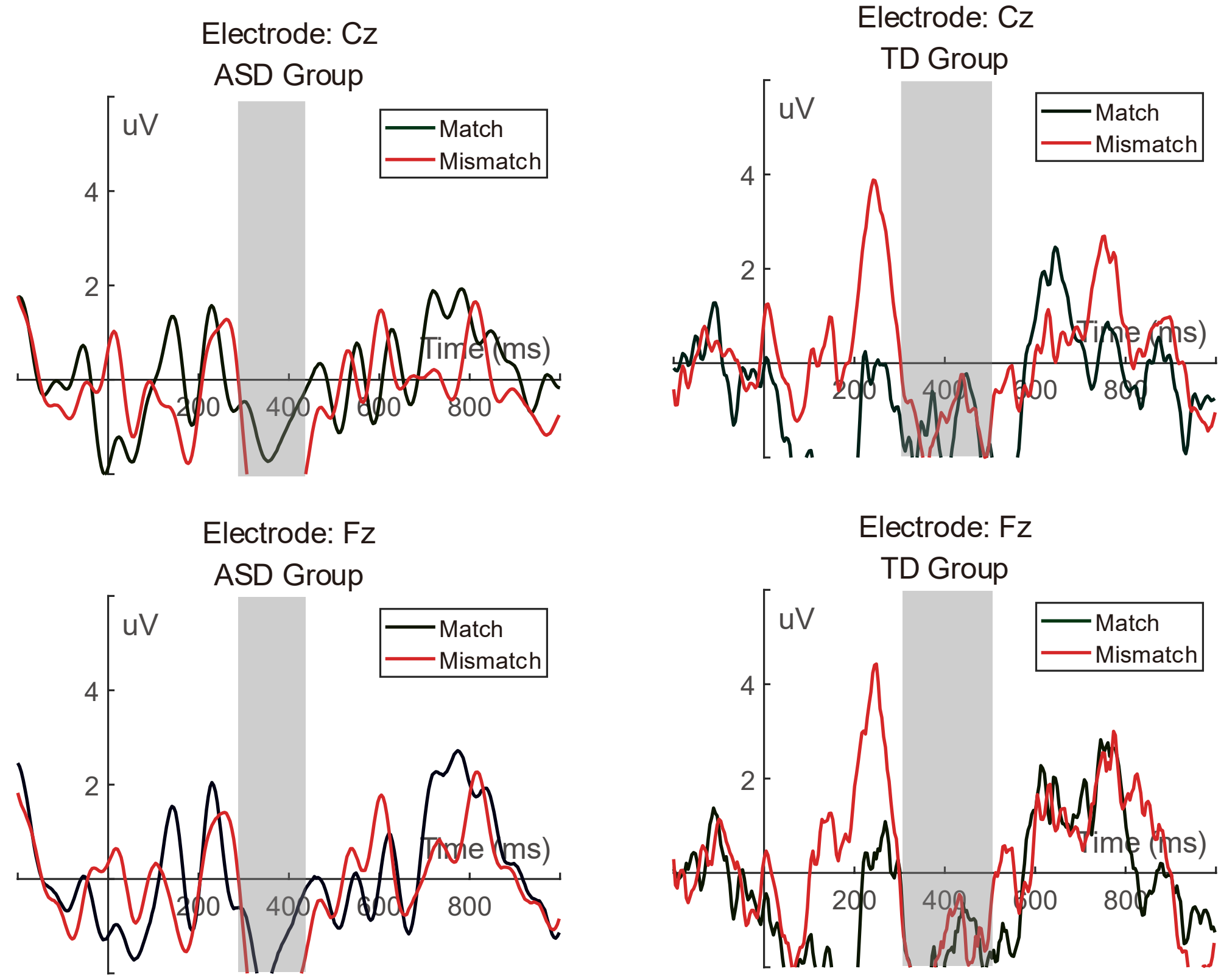

Figure 5 Grand Average ERP Waveforms at Midline Electrodes (Cz and Fz) in ASD and TD Groups

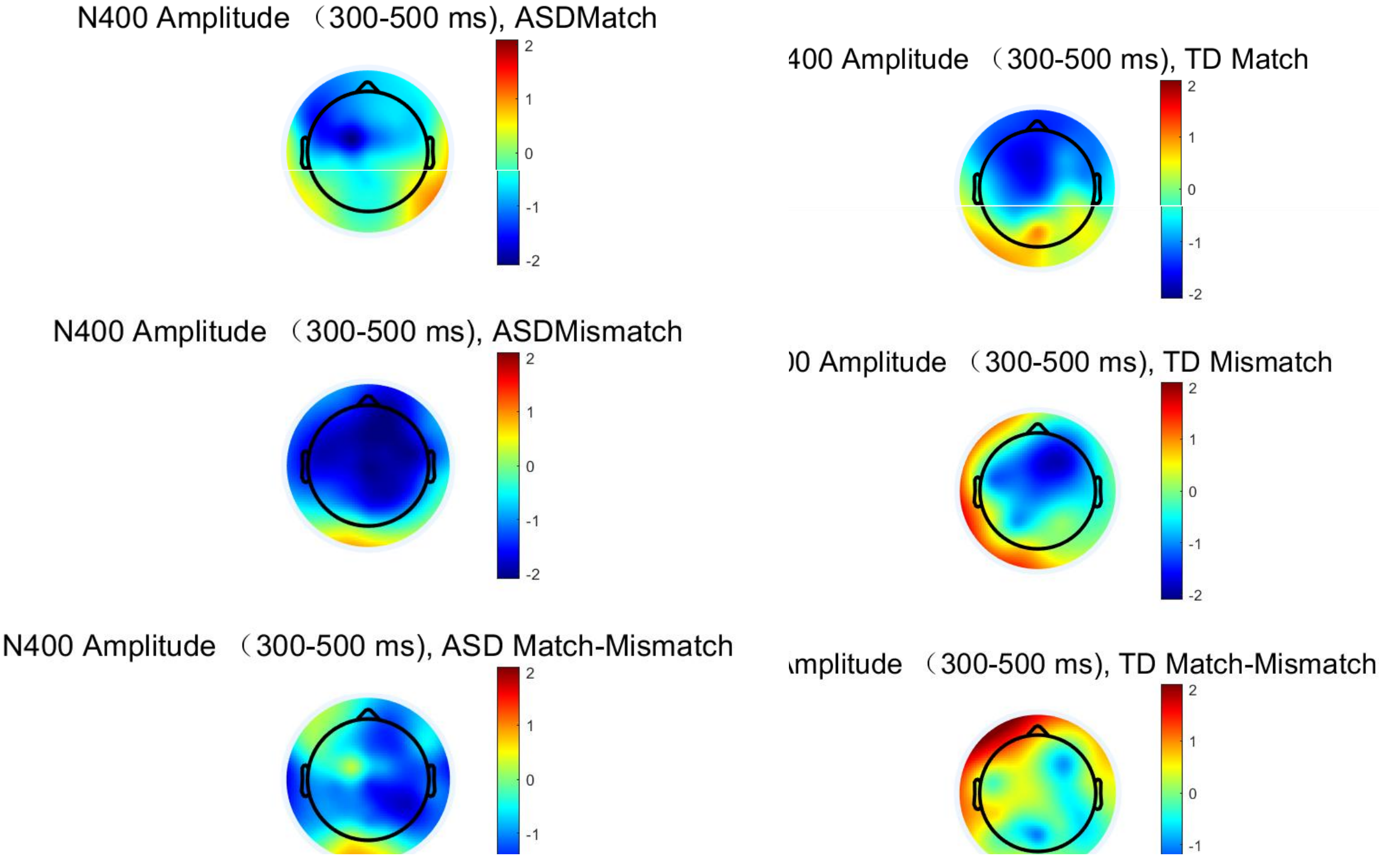

Figure 6 Scalp Topographies of Mean N400 Amplitudes for TD and ASD Group

Table 6 Descriptive Statistics of N400 Mean Amplitude (μV) for Each Group and Condition

| Group | Condition | M | SD |
|---|---|---|---|
| ASD | Match | -0.661 | 2.72 |
| | Mismatch | -1.59 | 1.65 |
| TD | Match | -0.732 | 2.73 |
| | Mismatch | -0.644 | 2.37 |

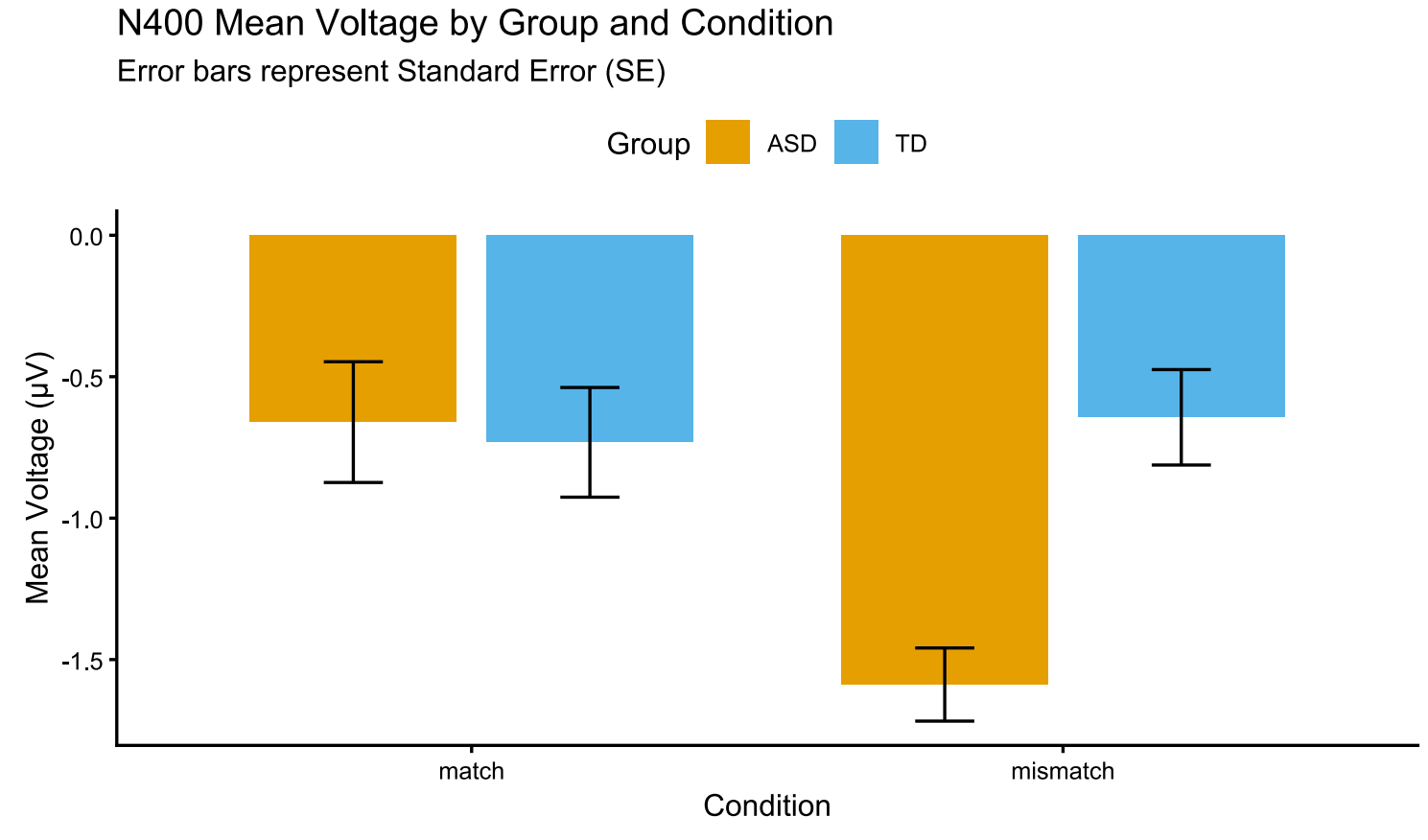

Figure 7 Interaction Effect of Group and Condition on N400 Mean Amplitude

Table 7 Type III ANOVA Results for the Linear Mixed-Effects Model of N400 Mean Amplitude

| Predictor | SS | MS | numdf | dendf | F | p |
|---|---|---|---|---|---|---|
| Group | 3.86 | 3.86 | 1 | 702.58 | 0.84 | 0.358 |
| Condition | 35.71 | 35.71 | 1 | 692.81 | 7.81 | .005** |
| Region | 111.43 | 37.14 | 3 | 692.81 | 8.12 | < .001*** |
| MA_c | 2.51 | 2.51 | 1 | 701.19 | 0.55 | 0.459 |
| Group × Condition | 46.24 | 46.24 | 1 | 692.81 | 10.11 | .002** |
| Group × Region | 9.63 | 3.21 | 3 | 692.81 | 0.7 | 0.551 |
| Condition × Region | 35.94 | 11.98 | 3 | 692.81 | 2.62 | .050* |
| Group × Condition × Region | 0.91 | 0.3 | 3 | 692.81 | 0.07 | 0.978 |

Table 8 Post-hoc Pairwise Comparisons of N400 Mean Amplitude Between ASD and TD Groups

| Condition | Contrast | Estimate | SE | df | t | p |
|---|---|---|---|---|---|---|
| Match | ASD – TD | 0.3 | 0.28 | 701 | 1.09 | 0.276 |
| Mismatch | ASD – TD | -0.72 | 0.28 | 701 | -2.59 | .010** |

### 4.2 RQ2: P600 (500–800 ms)

Mean amplitudes in the 500–800 ms window were analyzed using the same ROIs. Descriptive statistics showed that the TD group exhibited the canonical P600 effect, with larger amplitudes under mismatch (M = 1.06 μV, SD = 2.52) than match conditions (M = 0.47 μV, SD = 2.18). In contrast, the ASD group showed a reversed pattern (match: M = 0.78 μV, SD = 1.16; mismatch: M = 0.37 μV, SD = 1.22).

The LMM revealed no main effects of Group, $F(1, 699.12) = 1.29$, $p = .257$, or Condition, $F(1, 693.02) = 0.28$, $p = .595$, but a significant main effect of Region, $F(3, 693.02) = 4.91$, $p = .002$. Critically, the Group × Condition interaction was significant, $F(1, 693.02) = 14.21$, $p < .001$, indicating divergent condition effects across groups.

Post-hoc comparisons showed no group difference under match ($p = .207$), but significantly lower amplitudes in the ASD group under mismatch (estimate = −0.73 μV, $p = .002$). No higher-order interactions reached significance.

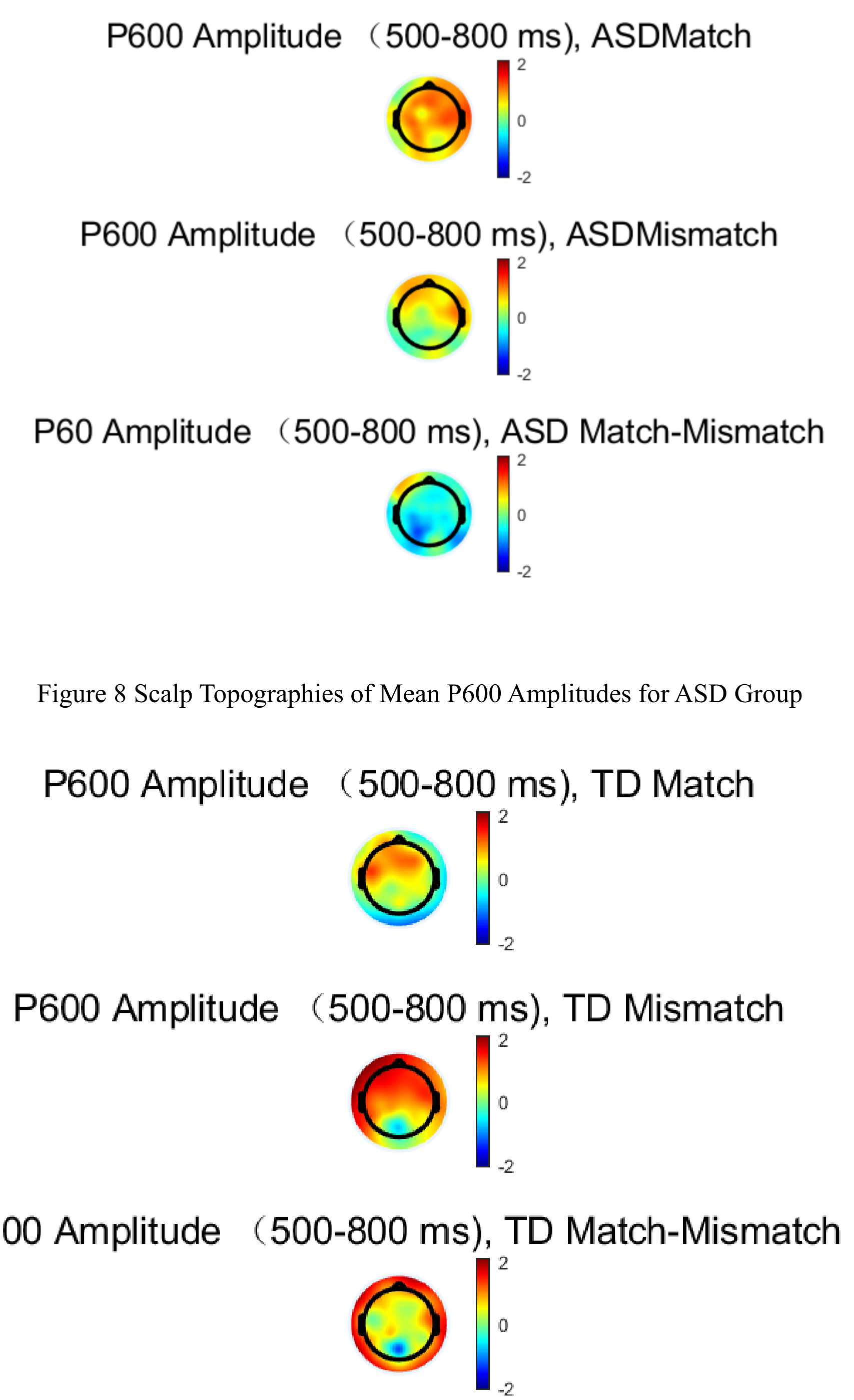

Figure 8 Scalp Topographies of Mean P600 Amplitudes for ASD Group

Figure 9 Scalp Topographies of Mean P600 Amplitudes for TD Group

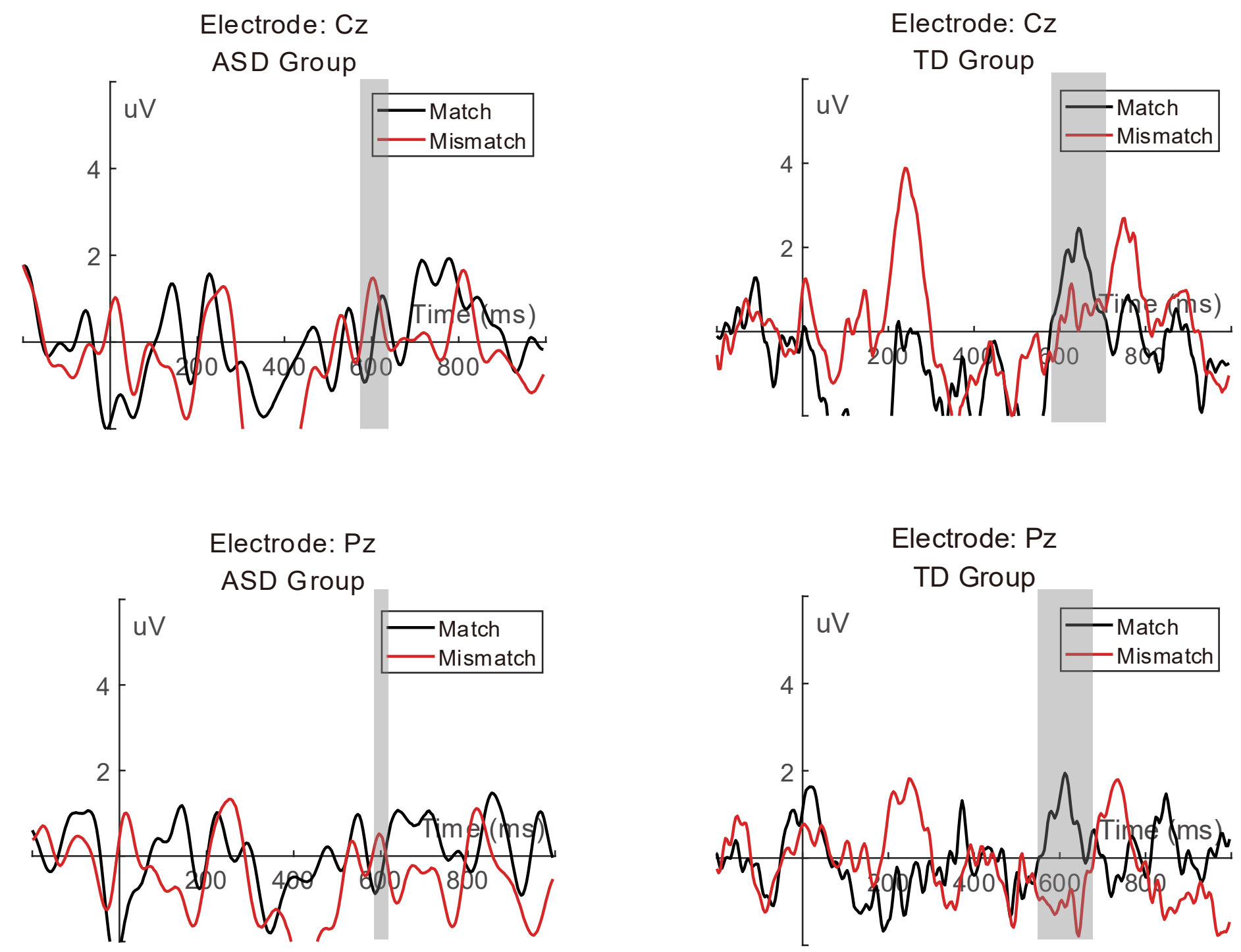

Figure 10 Grand Average ERP Waveforms for Match and Mismatch Conditions at Cz and PZ

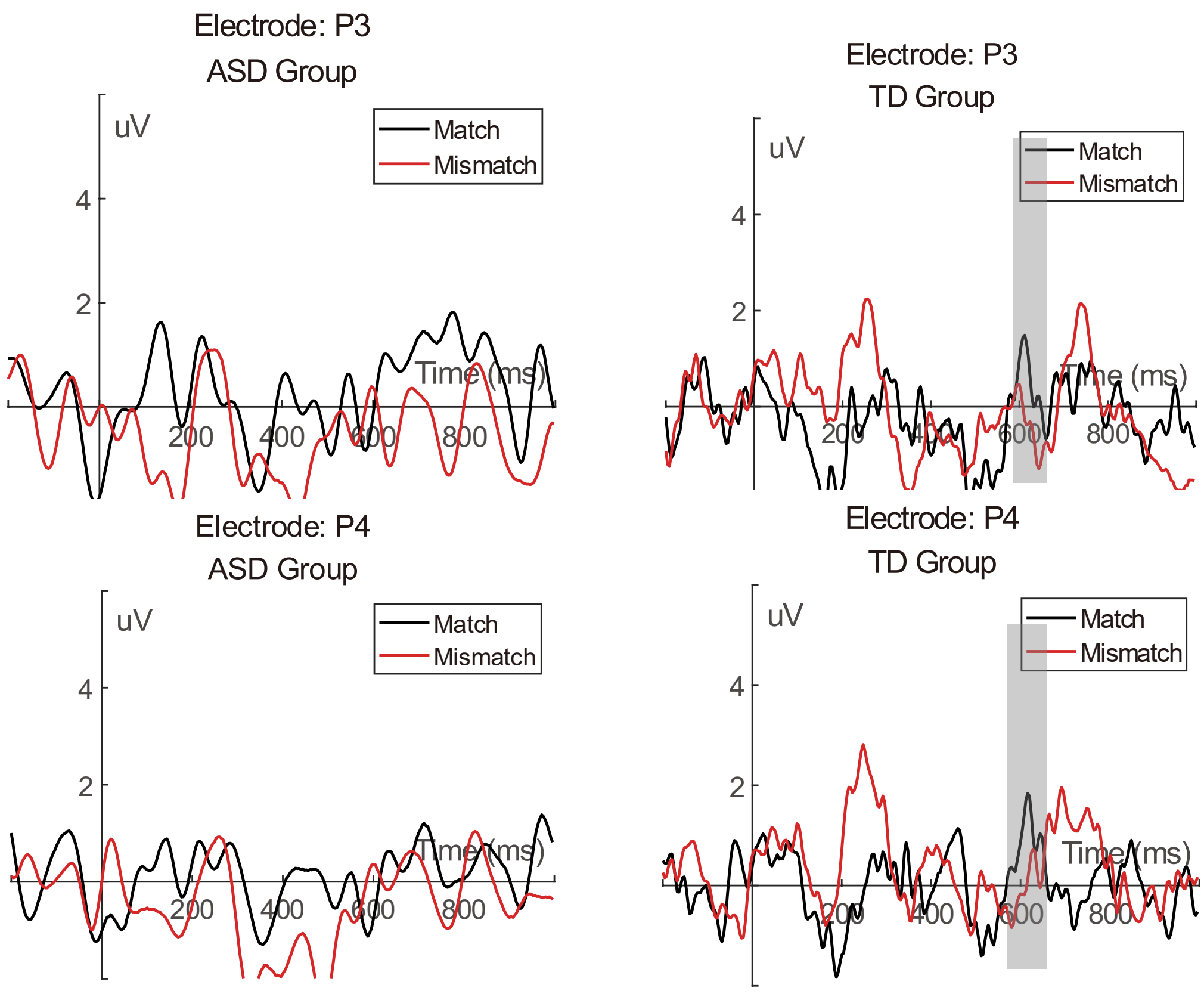

Figure 11 Grand Average ERP Waveforms for Match and Mismatch Conditions at P3 and P4

Table 9 Mean Voltages and Standard Deviations for Each Group and Condition

| Group | Condition | M | SD |
|---|---|---|---|
| ASD | Match | 0.78 | 1.16 |
| | Mismatch | 0.37 | 1.22 |
| TD | Match | 0.47 | 2.18 |
| | Mismatch | 1.06 | 2.52 |

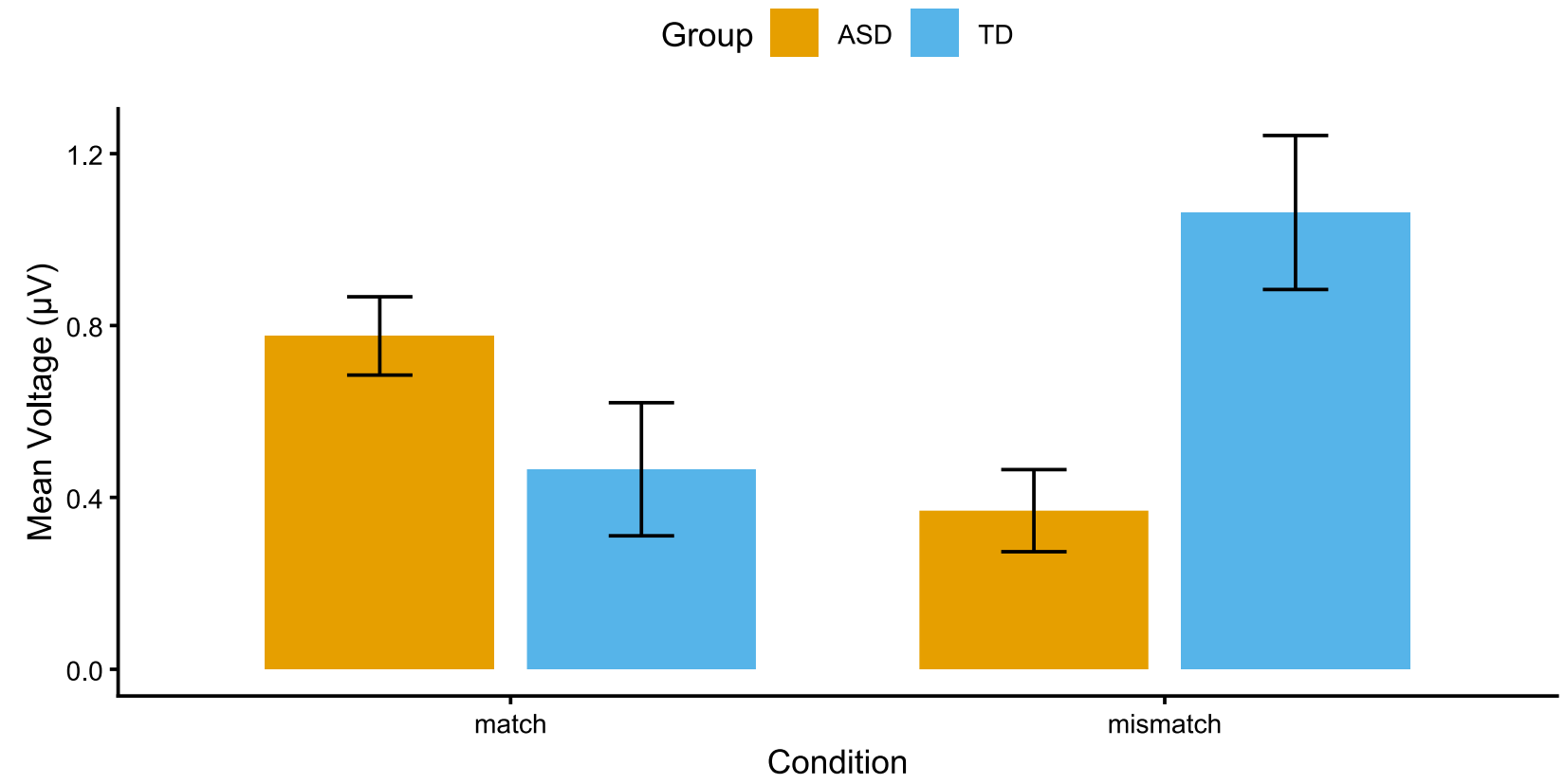

Figure 13 P600 Mean Voltage by Group and Condition

Table 10 Type III Analysis of Variance Table for the Linear Mixed Model (P600)

| Predictor | numdf | dendf | F | p |
|---|---|---|---|---|
| Group | 1 | 699.12 | 1.29 | 0.257 |
| Condition | 1 | 693.02 | 0.28 | 0.595 |
| Region | 3 | 693.02 | 4.91 | .002** |
| MA_c (Covariate) | 1 | 667.67 | 1.72 | 0.19 |
| Group × Condition | 1 | 693.02 | 14.21 | < .001*** |
| Group × Region | 3 | 693.02 | 1.32 | 0.267 |
| Condition × Region | 3 | 693.02 | 0.82 | 0.48 |
| Group × Condition × Region | 3 | 693.02 | 0.36 | 0.783 |

Table 11 Pairwise Comparisons of Group within Each Condition (P600)

| Condition | Comparison | Estimate | SE | p |
|---|---|---|---|---|
| Match | ASD – TD | 0.3 | 0.23 | 0.207 |
| Mismatch | ASD – TD | -0.73 | 0.23 | .002** |

## 4.3 RQ3: Lateralization Variability

Lateralization indices (LI) were computed based on P600 amplitudes. Absolute LI

values (|LI|) were used to assess variability. The ASD group showed significantly greater lateralization variability than the TD group, as indicated by a Wilcoxon rank-sum test (W = 119, p = .033) and a Fligner–Killeen test, $\chi^2(1) = 4.52$, p = .033.

Extreme lateralization cases (|LI| > Q3 + 1.5 × IQR) were significantly more frequent in the ASD group (Fisher's Exact Test, p = .024). However, robust regression analysis revealed no significant association between |LI| and PPVT scores within the ASD group (β = −52.56, p = .567), with negligible explained variance.

Table 12 Robust Regression Predicting PPVT from Absolute Lateralization in ASD Group

| Predictor | B | SE | t | p |
|---|---|---|---|---|
| Intercept | 66.86 | 20.69 | 3.23 | .005** |
| absLI | -52.56 | 89.86 | -0.585 | .567 |

*Note 1 N = 18 ASD participants. p < .01.*

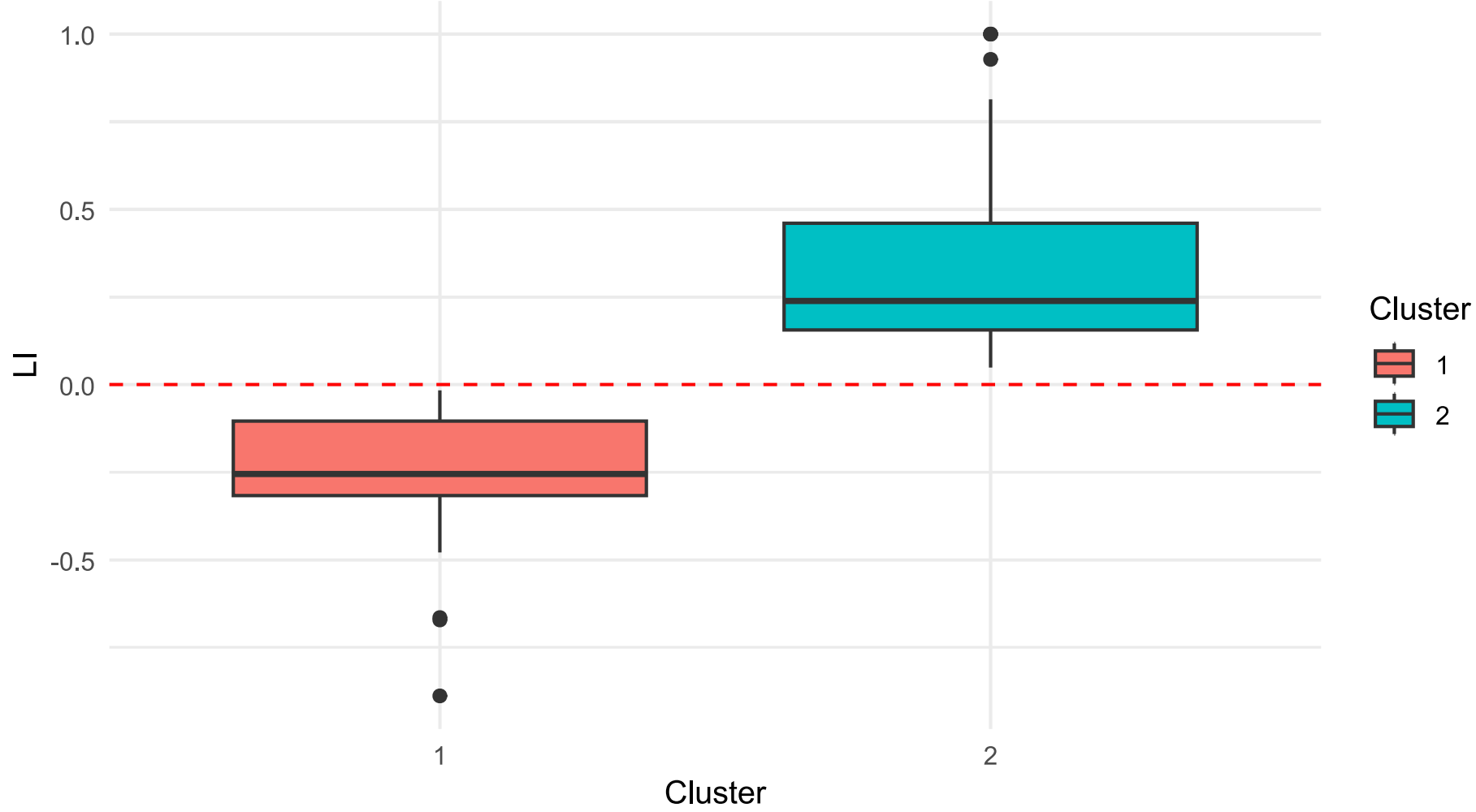

Figure 12 Boxplot Comparison of Lateralization Index (LI) by Cluster

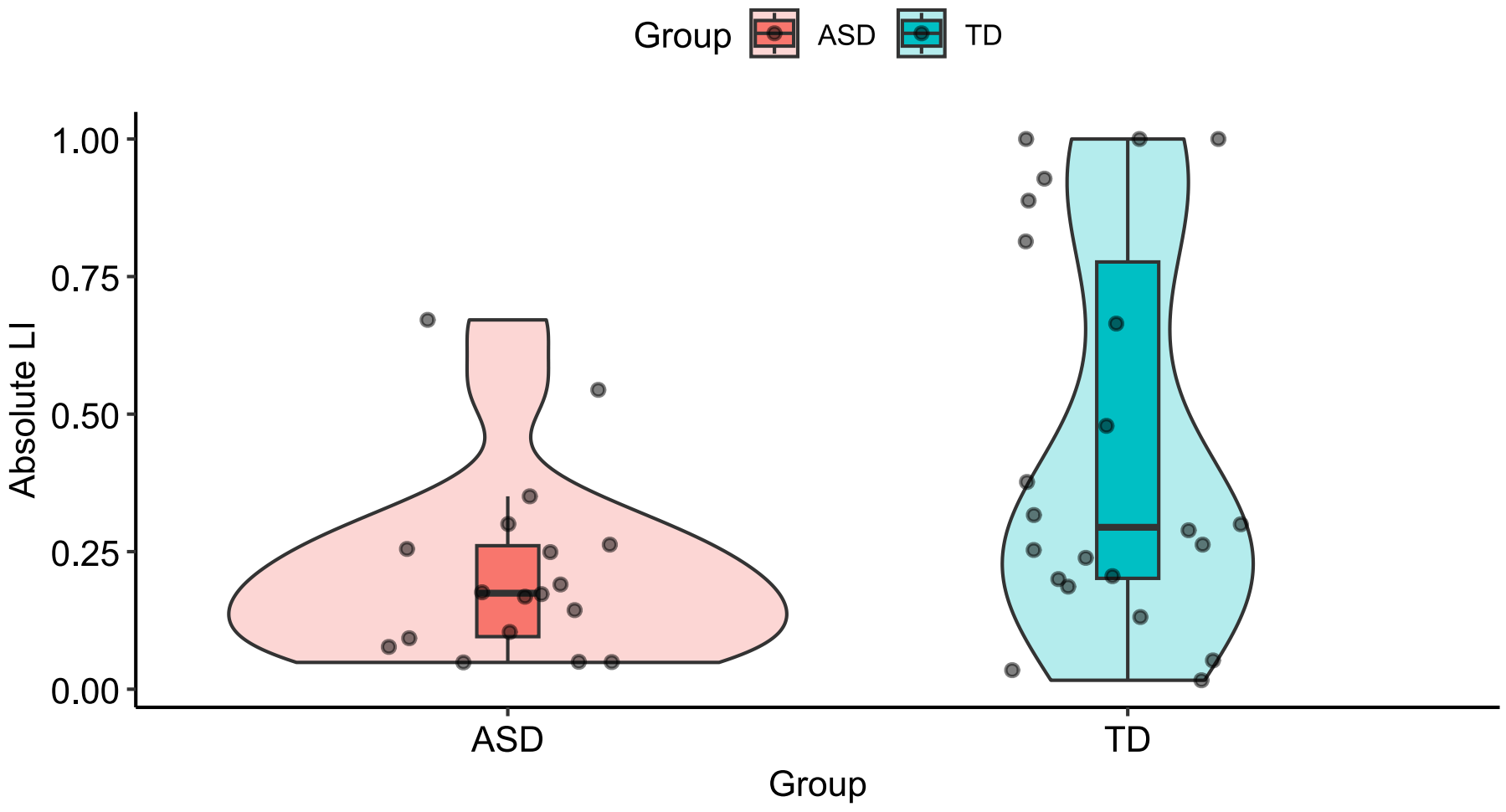

Figure 13 Violin Plot of Absolute Lateralization Index by Diagnostic Group

# 5. Discussion

## 5.1 Early Prediction and Semantic Integration

The absence of P200 modulation in the ASD group suggests impaired early-stage sensitivity to structural mismatch. In contrast, TD children showed enhanced P200 responses under mismatch, consistent with early predictive processing.

At the N400 stage, ASD children exhibited increased negativity under mismatch, indicating higher semantic integration costs. This pattern is consistent with a predictive processing account: reduced early prediction leads to greater reliance on later integration processes.

## 5.2 Late Syntactic Reanalysis

The P600 results indicate a qualitative group difference in late-stage processing. TD children showed a typical mismatch-related increase, reflecting syntactic reanalysis, whereas the ASD group did not.

This pattern supports a resource limitation account: earlier processing inefficiencies may reduce the resources available for late syntactic repair.

## 5.3 Lateralization Variability

ASD children exhibited greater inter-individual variability in hemispheric lateralization, including more extreme patterns. However, this variability was not associated with receptive vocabulary ability.

This dissociation suggests that neural organization during online syntactic processing may be more variable and less directly linked to standardized language measures.

## 6. Conclusion

The present study examined real-time processing of recursive locative structures in Mandarin-speaking children with ASD using ERP measures. Across three processing stages, the results revealed a consistent divergence between ASD and typically developing (TD) children.

At the early stage, TD children showed clear P200 modulation in response to structural mismatch, whereas ASD children exhibited no reliable differentiation, suggesting reduced engagement of early structural prediction. At the mid stage, ASD children showed enhanced N400 responses under mismatch conditions, indicating increased semantic integration demands. This pattern is consistent with a processing cascade in which reduced early prediction leads to greater reliance on later integrative processes.

At the late stage, TD children displayed the canonical P600 mismatch effect, reflecting syntactic reanalysis, while ASD children showed attenuated and non-differentiated responses. This suggests reduced recruitment of late-stage structural repair mechanisms, likely as a downstream consequence of earlier processing inefficiencies.

In addition, ASD children exhibited significantly greater variability in hemispheric lateralization during the P600 time window, with more frequent extreme lateralization patterns. However, lateralization magnitude was not significantly associated with receptive vocabulary performance, indicating that neural variability in recursive processing does not map directly onto standardized language measures.

Taken together, the findings support a cascading account of recursive processing in ASD, in which early predictive attenuation constrains later integration and reanalysis, alongside increased inter-individual variability in neural organization.

## 6.1 Implications

The present findings refine our understanding of recursive language processing in ASD at both syntactic and neurocognitive levels. Rather than indicating a deficit in recursive competence per se, the results suggest reduced efficiency in deploying recursive structure during online processing.

From a neurocognitive perspective, the data are consistent with reduced robustness of predictive mechanisms in ASD, particularly in morphosyntactic contexts involving recursive markers such as 的. The observed N400 and P600 patterns further indicate increased integration demands and reduced reanalysis efficiency during recursive structure building.

Importantly, these findings point to a difficulty at the interface between predictive processing and hierarchical integration, rather than a deficit localized to a single processing stage. This interpretation remains tentative and requires validation across different constructions, tasks, and languages.

Methodologically, the study demonstrates the feasibility of using a comprehension-based audiovisual paradigm to investigate recursive processing in pediatric ASD populations, while minimizing task-related artifacts and preserving sensitivity to hierarchical structure.

## 6.2 Limitations and Future Directions

Several limitations constrain the generalizability of the findings. First, the sample size was modest, particularly for analyses of lateralization variability, and the identified subgroups require replication in larger cohorts. Future studies should combine larger samples with more detailed cognitive profiling to clarify sources of inter-individual variability.

Second, the study examined only two-level recursive structures. Whether neural responses scale with increasing embedding depth remains an open question. Systematic manipulation of structural complexity would help determine how recursive load affects processing across stages.

Third, the exclusive use of auditory stimuli limits the ability to disentangle modality-specific effects from structural processing differences. Cross-modal designs would provide a more comprehensive account.

Fourth, the cross-sectional design precludes developmental interpretation. Longitudinal research is needed to determine whether the observed ERP patterns reflect delay, divergence, or stable differences in developmental trajectories.

Finally, the focus on locative recursion limits generalization across syntactic domains. Future work should examine multiple recursive constructions and extend investigation across languages to assess the generality of the observed effects.

More broadly, integrating ERP with complementary neuroimaging methods and cognitive measures will be essential for clarifying how predictive processing, hierarchical computation, and hemispheric organization jointly shape language processing in ASD.

Author contribution statement:

(Please note that one author may have more than one role or a job may have been done by multiple authors)

Wang Xiaoyi: making the EEG materials; collecting the EEG data; writing the whole paper; analyzing data; designing the experiment paradigm; recruiting participants; contacting laboratory equipment

Fu Chenxi: collecting the EEG data; designing the experiment paradigm; recruiting participants; contacting laboratory equipment

Ziman Zhuang: collecting the EEG data

Yang Caimei: supervising the whole EEG research group working at a series of theoretical and experimental studies related to recursion, part of which covers the present paper; supervise the whole research project; providing financial support for the project.

Funding acknowledgement: This work was supported by the National Social Science Fund of China (Grant Number: 23BYY170)